\RequirePackage{fix-cm}
\documentclass{svjour3}                     
\smartqed  

\usepackage{amssymb}
\usepackage{tabularx} 
\usepackage{graphicx}
\usepackage{algorithm, algorithmic}
\usepackage{amsmath}
\usepackage{color}
\usepackage{multirow}
\usepackage{url}
\usepackage{grffile}

\usepackage[table,xcdraw]{xcolor}

\begin{document}
	
	\title{Novel evaluation of surgical activity recognition models using task-based efficiency metrics}
	
	
	\author{Aneeq Zia\textsuperscript{1} \and Liheng Guo\textsuperscript{2} \and Linlin Zhou\textsuperscript{2} \and Irfan Essa\textsuperscript{1} \and Anthony Jarc\textsuperscript{2}
	}
	
	
	\institute{A. Zia \\
		\email{aneeqzia@gmail.com}\\
		\\
          \textsuperscript{1}College of Computing, Georgia Institute of Technology \\
North Ave NW, Atlanta, GA, USA 30332 \\
\textsuperscript{2}Medical Research, Intuitive Surgical, Inc. \\
5655 Spalding Drive, Norcross, GA, USA 30092
	}
	
	\date{Received: date / Accepted: date}

	\maketitle
\maketitle

\begin{abstract}

\textit{Purpose}: 
Surgical task-based metrics (rather than entire procedure metrics) can be used to improve surgeon training and, ultimately, patient care through focused training interventions. Machine learning models to automatically recognize individual tasks or activities are needed to overcome the otherwise manual effort of video review. Traditionally, these models have been evaluated using frame-level accuracy. Here, we propose evaluating surgical activity recognition models by their effect on task-based efficiency metrics. In this way, we can determine when models have achieved adequate performance for providing surgeon feedback via metrics from individual tasks.

\noindent \textit{Methods}:
We propose a new CNN-LSTM model, RP-Net-V2, to recognize the 12 steps of robotic-assisted radical prostatectomies (RARP). We evaluated our model both in terms of conventional methods (e.g. Jaccard Index, task boundary accuracy) as well as novel ways, such as the accuracy of efficiency metrics computed from instrument movements and system events.

\noindent \textit{Results}:
 Our proposed model achieves a Jaccard Index of 0.85 thereby outperforming previous models on robotic-assisted radical prostatectomies. Additionally, we show that metrics computed from tasks automatically identified using RP-Net-V2 correlate well with metrics from tasks labeled by clinical experts. 
 
\noindent \textit{Conclusions}: 
We demonstrate that metrics-based evaluation of surgical activity recognition models is a viable approach to determine when models can be used to quantify surgical efficiencies. We believe this approach and our results illustrate the potential for fully automated, post-operative efficiency reports.


\end{abstract}

\section{Introduction}

A primary goal of surgeons is to minimize adverse outcomes while successfully treating patients. Although many factors can influence outcomes, the technical skills of surgeons is one area shown to correlate \cite{birkmeyer2013surgical}. Therefore, methods to evaluate technical skills are critical. 

The most common approach for surgeon technical skill evaluation is expert feedback either intra-operatively in real-time or post-operatively through video review. However, an attending may not always be able to provide feedback in person, and post-operative video review can be time consuming and subjective. It is apparent this approach is not scalable, especially given the limited free time in expert surgeon schedules. Recently, crowd-sourced video evaluations have shown promise \cite{dai2017crowdsourcing}, but this approach still faces scalability and accuracy concerns. Automated, objective, and less time-consuming methods to evaluate surgeon technical skill are needed. 

Recently, objective, efficiency metrics derived from robotic-assisted surgical platforms have been defined for particular tasks within clinical procedures  \cite{chen2018objective}. They have been shown to differentiate expertise \cite{hung2018development} and to preliminary correlate to patient outcomes \cite{hung2018utilizing,hung2019deep}. These basic efficiency metrics closely resemble those from virtual reality simulators and stand to offer a scalable method for objective surgeon feedback. However, there remain two primary challenges for these objective metrics to impact surgery. The first challenge is to define which metrics matter most for individual surgical tasks. Efficiency metrics must be defined to control for a large amount of variability including anatomical variations and surgical approach or judgment. Such variation is not present in virtual reality tasks where these metrics are common since only a handful of standardized exercises is available \cite{liu2015review}. In pursuit of this first challenge, several clinical research teams are actively working to discover and validate efficiency metrics in clinical scenarios, as described earlier.  The second challenge is to automatically recognize the boundaries of surgical tasks. The begin and end times of tasks or sub-tasks must be automatically identified from within the entire procedure because manual identification through post-operative video review is overly time consuming and not scalable. Machine learning algorithms have been used with promising initial results in laparoscopic \cite{Padoy2012,Katić2015,Dergachyova2016,twinanda2017endonet} and robotic-assisted surgeries \cite{dipietro2016recognizing,ahmidi2017dataset,ahmidi2013string,gao2014jhu,lea2016segmental,zia_rarp_2018,zia2017temporal}. 

A critical shortcoming of the prior work in automatic, surgical activity recognition is that the models have been evaluated solely by their model accuracy but not their impact on task-based, efficiency metrics. For the application of advanced intra-operative or post-operative feedback through efficiency metrics, the requirements and specifications of machine learning models should be at least in part defined by their ability to result in accurate metrics. Certainly if the models perfectly predict every frame of a surgical procedure, they can also be used to perfectly compute efficiency metrics. However, most current models in literature range in accuracy from approximately 50$\%$ to 80$\%$ \cite{twinanda2017endonet,zia_rarp_2018}.

In this work, we define a new method to quantify the effects of surgical activity recognition models on technical, efficiency metrics during clinical tasks. In doing so, we work toward defining the necessary requirements for models used in automatic, surgeon performance reports.  

\emph{\textbf{Contributions:}} Our contributions are: (1) an improved model to automatically recognize the steps of clinical, robotic-assisted radical prostectomies and (2) a proposed, alternative method to evaluate models designed to recognize surgical activities based on their impact on task-based, surgeon efficiency metrics.

\begin{figure}[t]
	\centering
	\includegraphics[width=1.0\columnwidth]{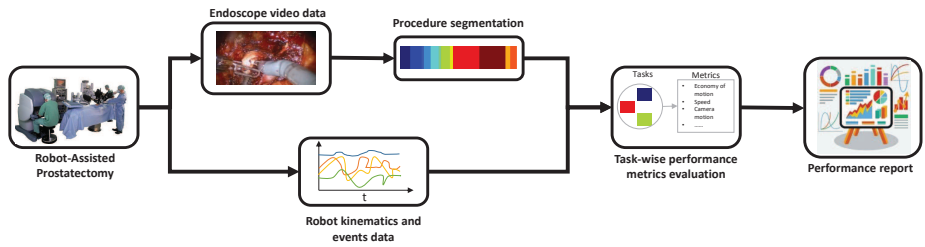}
	\caption{Flow diagram for automated performance report generation in robot-assisted prostatectomy where surgeons receive a readout of task-based metrics for steps of a surgery recognized using machine learning models.}
	\label{fig:flow_diagram}
	
\end{figure}

\section{Methodology}
The process to generate automated performance reports for robotic-assisted surgery is shown in Figure \ref{fig:flow_diagram}. First, procedure segmentation models divide a surgery into individual tasks. Next, the relevant efficiency metrics are computed using robotic system data. Finally, a report can be composed incorporating metrics for individual surgical tasks. This work focuses on the first two aspects leading up to the actual report and will not cover details of how such reports can be visualized. We will now provide more details regarding the first two parts of the pipeline below.

\begin{figure}[t]
	\centering
	\includegraphics[width=1.0\columnwidth]{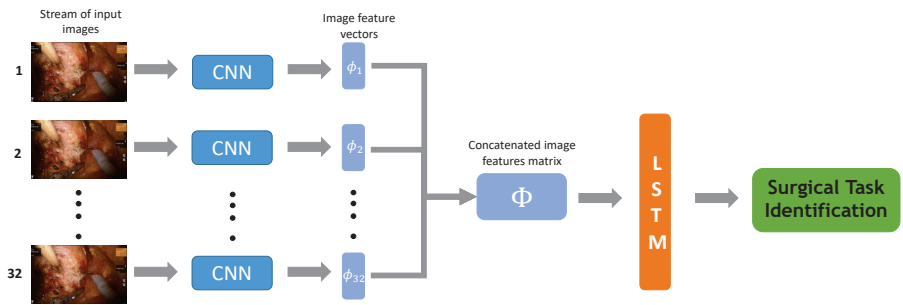}
	\caption{\textit{`RP-Net-V2'} for automatic procedure segmentation. A stream of input images are passed through individual convolutional neural networks to extract image feature vectors which are then passed through a long short-term memory network to identify surgical tasks.}
	\label{fig:model}
	
\end{figure}

\noindent\emph{\textbf{Procedure Segmentation:}} 
Previously, \cite{zia_rarp_2018} presented a detailed comparison of single image-based and kinematics-based models for this task. Their results showed that single image based models perform significantly better than models based solely on robotic system data. While the results of using single image based models were promising, the work presented in \cite{zia_rarp_2018} did not sufficiently utilize temporal information, for example by using streams of images in order to extract motion based features from videos. Some of the recent works in the field \cite{kannan2018future,sarikaya2018joint,endo3d,jin2018sv} have explored various model architectures that try to learn visual and motion features together from a stream of images. Similar to such works, in this paper, we use a CNN-LSTM based model `RP-Net-V2' for robotic-assisted radical prostatectomy activity recognition as shown in Figure \ref{fig:model}. Our model takes in a stream of N consecutive images at 1 fps. Each image is input into a separate Convolutional Neural Network (CNN) to learn visual features from all input images. The CNN models have shared weights. Outputs of all CNN models are then stacked together to produce a feature matrix  $\Phi \in \Re^{F \times N}$, where $F$ is the dimension of output feature vector from the CNN models. The matrix $\Phi$ represents the visual feature information from a stream of images which is then used as input into a LSTM network in order to extract temporal information from the visual feature vectors. The LSTM network then outputs the predicted surgical task. Training of this network is done in an end-to-end manner using stochastic gradient descent (SGD) with a categorical cross-entropy loss. 

We also employ a model for utilizing image and system data together for procedure segmentation. For this, the same architecture as that described for RP-Net-V2 is used, however,  along with image frames, windows of system data coming from different streams are also fed into the model. Instead of a CNN, an LSTM network is used to extract features from the system data branches before concatenating all image based and system data based features together.

Similar to \cite{zia_rarp_2018}, we also employ a running window median filter of length $W$ on the procedure segmentation output to remove unwanted misclassification spikes. Despite post-processing, mis-classifications can still exist causing cases where the predictions have disconnected continuous segments of the same task. Therefore, for each task, we select the longest continuous segment as our model's predicted task annotation. The extracted tasks are then used for performance metrics evaluation.

\noindent\emph{\textbf{Efficiency Metrics Computation:}}
We evaluate efficiency metrics in two ways. The first involved taking the longest continuous segment predicted per task from the model and evaluating metrics only on that segment. In the second approach, we evaluated different metrics on all discontinuous predicted segments and combined them to get the final metric value. The metrics are computed from robotic system data, such as joint angles (or kinematics) and button presses (events), specific to the da Vinci surgical system (details of dataset given in experimental evaluation section). The metrics used in this work are the same as those previously shown to be useful in differentiating surgeon experience and correlating to outcomes \cite{hung2018development,hung2018utilizing}. We evaluated a total of 13 kinematics based and 33 events based metrics. Note these totals include the same metrics evaluated using different parts of the system. For example, economy of motion was evaluated for both patient side manipulators making it 2 of the 14 total number of metrics. Similarly for events based metrics, camera control on and off are treated as separate metrics. All the event-based metrics are discrete in nature while the kinematic-based metrics are continuous. Some examples of these metrics are given in Table \ref{table_metrics}.

\noindent\emph{\textbf{Efficiency Metrics Evaluation:}} In order to evaluate how well our proposed RP-Net-V2 works for metrics computation, we compare the resulting efficiency metrics with those from human task labels (i.e., ground truth). Fo this, we used the Pearson's correlation coefficient between ground truth and model predicted metrics to evaluate performance.

\begin{table}[b]
	\caption{Few examples of events and kinematics based metrics used for evaluation.}
	\resizebox{\textwidth}{!}{%
		\begin{tabular}{|l|l|}
			\hline
			Data Class   &   Names of metrics used \\ \hline
			Event based      &       Camera control on/off, energy on/off,  hand controller clutch on/off, head in/out, arm swap, etc.               \\ \hline
			Kinematics based &        Economy of motion, wrist angles (roll, pitch, yaw), speed, etc.                \\ \hline
		\end{tabular}
	}
	\label{table_metrics}
\end{table}

\section{Experimental Evaluation}
\textbf{Dataset:} 
We had access to the data set from \cite{zia_rarp_2018} and use that for our model training and evaluations. The data set consists of 100 prostatectomies from a single center with 12 tasks each. The 12 tasks are listed in Table \ref{table_tasks}. System data was collected at 50Hz and synchronized to a single channel of endoscopic video captured at 30 fps. A single, surgical resident manually identified the begin and end times for each surgical task through post-operative video review. Similar to \cite{zia_rarp_2018}, we divide the dataset into three sets: 70 procedures are used for training, 10 for validation, and the remaining 20 procedures are used to test model performances. All results presented in the next section are evaluated on the test set.

\begin{table}[t]
	\centering
	\caption{Dataset: the 12 steps of robot-assisted radical prostatectomy.}
	\label{table_tasks}
	\begin{tabular}{|c|c|c|c|}
		\hline
		Task no & Task Name                          \\ \hline
		1       & mobilize colon / drop bladder            \\ \hline
		2       & Endopelvic fascia / DVC                     \\ \hline
		3       & Anterior bladder neck dissection       \\ \hline
		4       & Posterior bladder neck dissection              \\ \hline
		5       & Seminal vesicles                          \\ \hline
		6       & Posterior plane / Denonvilliers                 \\ \hline
		7       & Predicles / nerve sparing                     \\ \hline
		8       & Apical dissection                           \\ \hline
		9       & Posterior anastomosis                        \\ \hline
		10      & Anterior anastomosis                            \\ \hline
		11      & Lymph node dissection L                        \\ \hline
		12      & Lymph node dissection R                          \\ \hline
	\end{tabular}
\end{table}

\noindent\textbf{Parameter estimation:}
There are many parameters that we need to tune in order to achieve optimal performance. We use validation accuracies for all parameter selection. The proposed procedure segmentation model `RP-Net-V2' takes a window of $N$ consecutive images at 1 fps. Using a low value of $N$ would result in the model having less temporal information but with more overall training data, and vice versa. We also need to specify how frequent each window of images is sampled and define $O$ as the fraction of images overlapping between two consecutive windows of training samples. We run our experiments for $N \in [2,4,8,16,32,64]$ and $O \in [0,0.5]$ with all possible combinations between $N$ and $O$, and selected $N = 32$ with $O = 0.5$ based on highest validation accuracy. Moreover, the dimensions of the visual feature vector extracted using CNN models (represented by $F$) can also affect performance. We used values of $F$ ranging from 64 to 2048 and selected $F=1024$. 

For the CNN models, we experiment with most of the popular architectures like VGG-16, VGG-19, Inception-V3 and ResNet-50. Both randomly initialized weights and pre-trained weights from ImageNet \cite{deng2009imagenet} were experimented with. The best validation accuracy was achieved when using VGG-19 with ImageNet pre-trained weights. For the RNN network, we implemented single layered bi-directional models with LSTM units ranging from 32 till 1024, and achieved the best performance when using 256 units. In post-processing, we use the same window length ($W = 301$) for the median filter as presented in \cite{zia_rarp_2018}. It is important to note that the window length used is specific to our dataset and will have to be re-evaluated when using on a different surgery e.g any procedure with much shorter or longer tasks as compared to prostatectomy.

\section{Results and Discussion}

\begin{figure}[t]
	\centering
	\includegraphics[width=1.0\columnwidth]{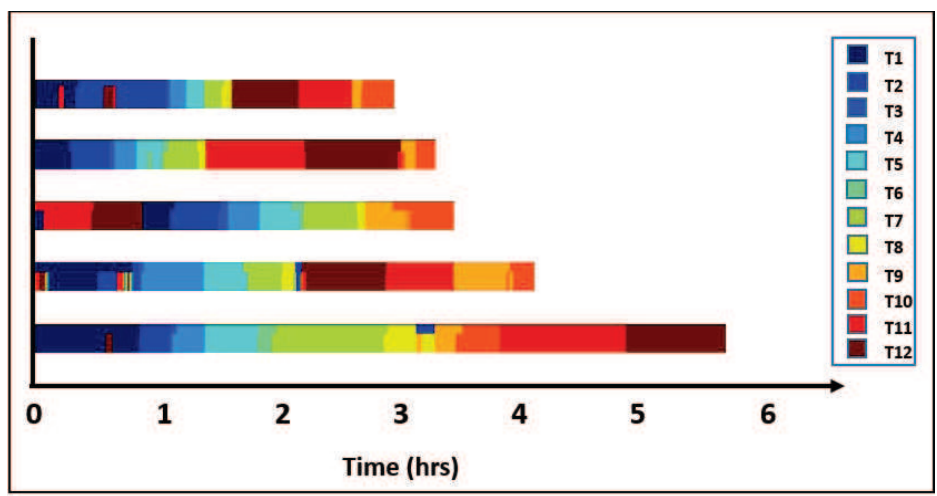}
	\caption{Procedure segmentation outputs from RP-Net-V2 after post-processing. Each bar represents one whole procedure in the test set. Within each bar, the top half shows the ground truth labels while the lower half shows predictions. Color coding for the different tasks is also shown. (Best viewed in color)}
	\label{fig:segmentation_bars}
	
\end{figure}

Our results showed that the proposed RP-Net-V2 outperformed previous models both before and after post-processing as shown in Table \ref{table:results_accuracies}. This highlights the importance and value of temporal information captured in a set of subsequent image frames. Additionally, VGG-19 outperformed other conventional models as the foundation for RP-Net-V2. We also check for statistical significance for improvements in classification using Mcnemar's test \cite{McNemar1947}. Comparing the performance of RP-Net-V2 versus RP-Net, we found that the improvements achieved by RP-Net-V2 in recognizing surgical activities are statistically significant with a $p-value<0.01$. Moreover, feeding system data into the RP-Net-V2 architecture along with images in a multi-modal fashion deteriorated the performance. This can be explained by the fact that system data models were shown to perform poorly in \cite{zia_rarp_2018} - hence adding that information in the model had a negative impact on the final model performance. Qualitative examples of procedure bars after filtering are shown in Figure \ref{fig:segmentation_bars} when using the best model (RP-Net-V2-VGG19). Similar to \cite{zia_rarp_2018}, a simple post-processing median filter removed many misclassified frames and short segments. However, such a filter is only possible for post-operative reports since it is not a real-time estimate. 

\begin{table}[t]
	\centering
	\label{table:results_accuracies}
	\caption{Table of results showing average Jaccard Index along with standard deviations when using different models.}
	\begin{tabular}{|l|l|l|}
		\hline
		& Unfiltered                                & Filtered                                  \\ \hline
		LSTM[ssc+evt](MS) \cite{zia_rarp_2018} & 0.63 $\pm$ 0.18                           & 0.66 $\pm$ 0.19                           \\ \hline
		RP-Net \cite{zia_rarp_2018}                        & 0.70 $\pm$ 0.05                           & 0.81 $\pm$ 0.07                           \\ \hline
		RP-Net-V2 (VGG19) + LSTM[ssc+si+evt](MS)                     & 0.71 $\pm$ 0.05                           & 0.74 $\pm$ 0.06                           \\ \hline
		RP-Net-V2 (InceptionV3)                                               & 0.79 $\pm$ 0.04                           & 0.83 $\pm$ 0.05                           \\ \hline
		\textbf{RP-Net-V2 (VGG19)}                         & \textbf{0.81 $\pm$ 0.06} & \textbf{0.85 $\pm$ 0.07} \\ \hline
	\end{tabular}
\end{table}

\begin{figure}[t]
	\centering
	\includegraphics[width=0.9\columnwidth]{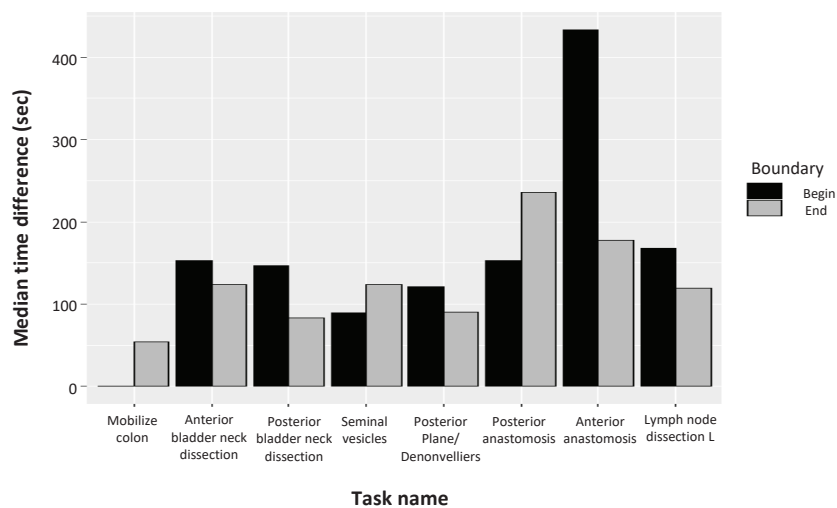}
	\caption{Median difference (in seconds) between human labeled and predicted task boundaries for both task beginning and end.}
	\label{fig:boundary_median_error}
	
\end{figure}

\begin{figure}[t]
	\centering
	\includegraphics[width=1.0\columnwidth]{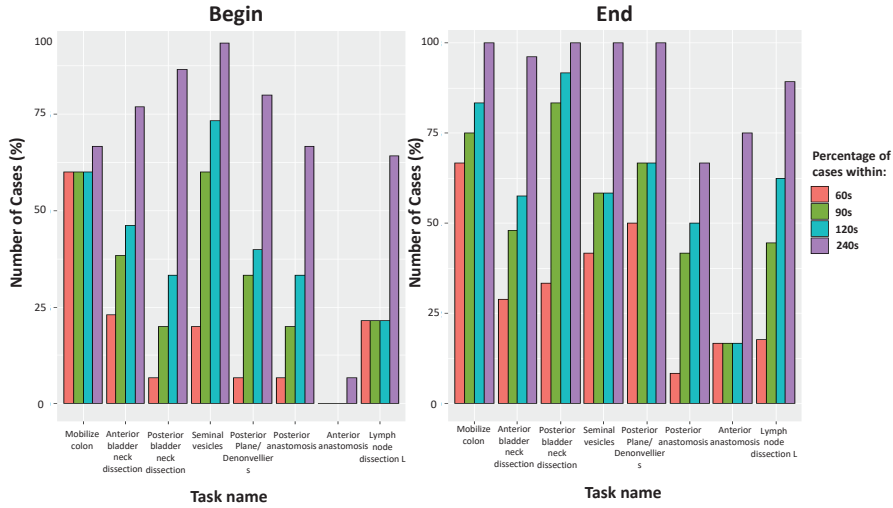}
	\caption{Bar graphs showing how far predicted begin and end times are from ground truth for different tasks. Different colored bard for each task represents a different range of time difference. (Best viewed in color)}
	\label{fig:boundary_percentage}
\end{figure}

In order to see how our model predicted task boundaries compare to ground truth in terms of time, we show median errors (in seconds) for the predicted begin and end boundaries of some surgical tasks in Figure \ref{fig:boundary_median_error}. We also analyze the percentage of cases for which individual tasks predicted begin/end times lie within a specific range as shown in Figure \ref{fig:boundary_percentage}. From the two figures, we can see that certain tasks have quite small errors in boundary predictions whereas other tasks like T10 (Anterior anastomosis) have much larger errors.  This could be a result of these different tasks having different variability across patients and technique. Some tasks might be very standardized with minimal anatomical variations whereas others require different approaches each time due to patient differences. However, even with these errors in task boundary predictions, we can see from Figure \ref{fig:boundary_percentage} that for the majority of tasks, more than $80\%$ of the cases lie within 4 minutes ($240s$) of ground truth. Considering the average overall duration of the prostatectomies in our dataset was 122.2 minutes, such an error does not seem large.

\begin{table}[]
	\centering
	\caption{Results for human annotation vs model annotation metrics evaluation comparison when using \textbf{longest predicted segments}. Each cell shows the average pearson's correlation coefficient achieved over all metrics evaluated for the corresponding task for the two models in the format \textit{RP-Net $\vert$ RP-Net-V2}. We use the median $p-value$ across all metrics for each task to represent statistical significance - a * next to correlation average values represents median $p-value < 0.05$. }
	\resizebox{\textwidth}{!}{%
	\begin{tabular}{|l|l|l|}
		\hline
		Task                              & Average Correlation Coefficient (EVT)     & Average Correlation Coefficient (KIN)    \\ \hline
		mobilize colon / drop bladder     & 0.70 $\pm$ 0.20* $\vert$ 0.88 $\pm$ 0.15* & 0.75 $\pm$ 0.26* $\vert$ 0.92 $\pm$ 0.09* \\ \hline
		Anterior bladder neck dissection  & 0.43 $\pm$ 0.26 $\vert$ 0.81 $\pm$ 0.27*  & 0.46 $\pm$ 0.37* $\vert$ 0.59 $\pm$ 0.24* \\ \hline
		Posterior bladder neck dissection & 0.41 $\pm$ 0.36 $\vert$ 0.73 $\pm$ 0.22*  & 0.53 $\pm$ 0.35* $\vert$ 0.70 $\pm$ 0.26* \\ \hline
		Seminal vesicles                  & 0.59 $\pm$ 0.29* $\vert$ 0.87 $\pm$ 0.11* & 0.73 $\pm$ 0.20* $\vert$ 0.93 $\pm$ 0.04* \\ \hline
		Posterior plane / Denonvilliers   & 0.75 $\pm$ 0.22* $\vert$ 0.41 $\pm$ 0.41  & 0.39 $\pm$ 0.24 $\vert$ 0.50 $\pm$ 0.32  \\ \hline
		Apical dissection                 & 0.62 $\pm$ 0.16* $\vert$ 0.75 $\pm$ 0.08* & 0.79 $\pm$ 0.17* $\vert$ 0.78 $\pm$ 0.20* \\ \hline
		Posterior anastomosis             & 0.54 $\pm$ 0.34* $\vert$ 0.46 $\pm$ 0.34  & 0.53 $\pm$ 0.42* $\vert$ 0.47 $\pm$ 0.37* \\ \hline
		Anterior anastomosis              & 0.60 $\pm$ 0.13* $\vert$ 0.53 $\pm$ 0.31* & 0.28 $\pm$ 0.26 $\vert$ 0.70 $\pm$ 0.24* \\ \hline
		Lymph node dissection Left        & 0.84 $\pm$ 0.08* $\vert$ 0.93 $\pm$ 0.03* & 0.70 $\pm$ 0.25* $\vert$ 0.79 $\pm$ 0.18* \\ \hline
		\end{tabular}
}
\label{table:results_metrics}
\end{table}

\begin{table}[]
	\centering
	\caption{Results for human annotation vs model annotation metrics evaluation comparison when using \textbf{all predicted segments}. Each cell shows the average pearson's correlation coefficient achieved over all metrics evaluated for the corresponding task for the two models in the format \textit{RP-Net $\vert$ RP-Net-V2}. We use the median $p-value$ across all metrics for each task to represent statistical significance - a * next to correlation average values represents median $p-value < 0.05$. }
	\resizebox{\textwidth}{!}{%
		\begin{tabular}{|l|l|l|}
		\hline
		Task                              & Average Correlation Coefficient (EVT)     & Average Correlation Coefficient (KIN)     \\ \hline
		mobilize colon / drop bladder     & 0.76 $\pm$ 0.14* $\vert$ 0.89 $\pm$ 0.22* & 0.69 $\pm$ 0.30* $\vert$ 0.82 $\pm$ 0.19* \\ \hline
		Anterior bladder neck dissection  & 0.14 $\pm$ 0.26 $\vert$ 0.82 $\pm$ 0.27*  & 0.28 $\pm$ 0.40 $\vert$ 0.53 $\pm$ 0.24*  \\ \hline
		Posterior bladder neck dissection & 0.66 $\pm$ 0.43 $\vert$ 0.91 $\pm$ 0.06*  & 0.37 $\pm$ 0.28 $\vert$ 0.60 $\pm$ 0.31*  \\ \hline
		Seminal vesicles                  & 0.84 $\pm$ 0.09* $\vert$ 0.77 $\pm$ 0.13* & 0.35 $\pm$ 0.27 $\vert$ 0.68 $\pm$ 0.23*  \\ \hline
		Posterior plane / Denonvilliers   & 0.62 $\pm$ 0.05* $\vert$ 0.41 $\pm$ 0.38  & 0.20 $\pm$ 0.18 $\vert$ 0.45 $\pm$ 0.34   \\ \hline
		Apical dissection                 & 0.71 $\pm$ 0.12* $\vert$ 0.74 $\pm$ 0.29* & 0.25 $\pm$ 0.32 $\vert$ 0.36 $\pm$ 0.34   \\ \hline
		Posterior anastomosis             & -0.23 $\pm$ 0.24 $\vert$ 0.34 $\pm$ 0.51  & 0.10 $\pm$ 0.35 $\vert$ 0.40 $\pm$ 0.33*  \\ \hline
		Anterior anastomosis              & -0.37 $\pm$ 0.02 $\vert$ 0.30 $\pm$ 0.20  & 0.12 $\pm$ 0.34 $\vert$ 0.42 $\pm$ 0.39*  \\ \hline
		Lymph node dissection Left        & 0.93 $\pm$ 0.03* $\vert$ 0.88 $\pm$ 0.05* & 0.43 $\pm$ 0.18 $\vert$ 0.56 $\pm$ 0.19*  \\ \hline
	\end{tabular}
}
\label{table:results_metrics_discontinuous}
\end{table}

\begin{figure}[t]
	\centering
	\includegraphics[width=1.0\columnwidth]{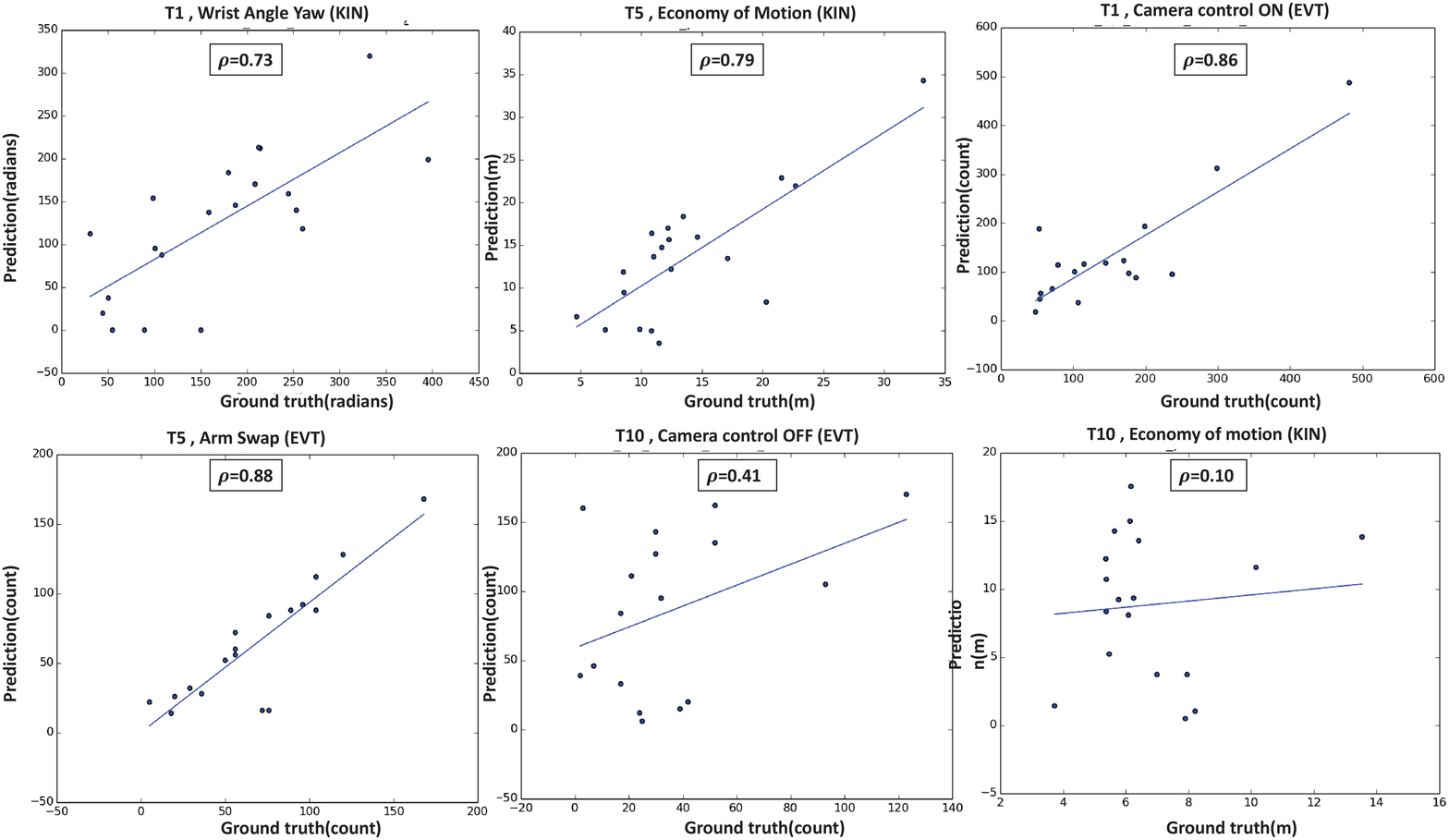}
	\caption{Scatter plots performance metrics evaluated using model annotations vs human annotations. Task number, metric type and correlation coefficient value is also shown for each plot. The right two plots on the bottom row show some bad cases while the rest show good cases.}
	\label{fig:metrics}
	
\end{figure}

Looking at final performance metrics predictions, our model based annotations resulted in certain tasks performing very well (e.g Seminal vesicles, Lymph node dissection), whereas some did not (e.g Posterior anastamosis) as can be seen in Table \ref{table:results_metrics} and \ref{table:results_metrics_discontinuous}. We also tested statistical significance of our results using a two-tailed test and incorporated those in the tables. In general, event based metrics might be more sensitive to inaccuracies on begin/end boundaries than kinematic since events come in bursts - if you miss a burst, it can significantly influence the metric. Example scatter plots for individual metrics are shown in Figure \ref{fig:metrics}. In an ideal case, the values would fall on the unity line. Again, certain metrics follow this trend more than others. In general, \textit{RP-Net-V2} achieves significance on several more tasks than \textit{RP-Net}, and the correlation coefficients are generally greater for \textit{RP-Net-V2} than \textit{RP-Net}. Additionally, this trend holds true for both metrics computed from the longest predicted segments (\ref{table:results_metrics}) as well as all predicted segments (\ref{table:results_metrics_discontinuous}).

Importantly, not all tasks have the same impact on surgeon training and outcomes. For example, there may be critical steps that are solely responsible for certain outcomes (e.g., cutting a nerve during a nerve dissection). Similarly, there are other tasks that are critical for training new surgeons by introducing them to both technical and cognitive skills. Based on \cite{hung2018experts}, the cardinal steps of prostatectomy include bladder neck dissection, apical dissection and anastomosis. These were selected based on their impact to both outcomes and surgeon training. From Tables \ref{table:results_metrics} and \ref{table:results_metrics_discontinuous}, we see a reasonably positive average correlation values for these tasks illustrating that automated performance reports are feasible for several of the most clinically relevant tasks.

\begin{figure}[t]
	\centering
	\includegraphics[width=1.0\columnwidth]{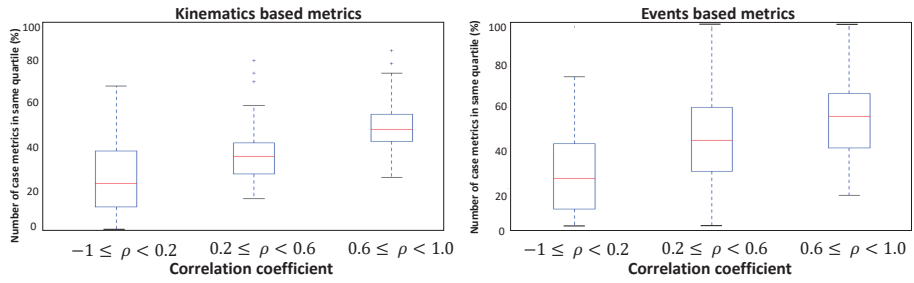}
	\caption{Box plots comparing correlation coefficients with number of surgeon's lying in the same quartile of predicted and ground truth metrics.}
	\label{fig:metrics_boxplot}
\end{figure}

While we do not perform too well on some tasks in terms of metrics predictions, this may not be that critical in terms of giving performance based feedback. Surgeons may not always want to know their exact performance metrics on various tasks and might just want to see how their numbers compare to their peers. For this, we performed another experiment to see that for a specific task, does a surgeon's predicted performance metric lies in the same quartile of all predictions when compared to ground truth. Figure \ref{fig:metrics_boxplot} shows how correlation coefficient relates to number of surgeon's lying in the same quartile of predictions and ground truth. As expected, a higher correlation coefficient would result in a higher number of surgeon's lying in the same quartile since our predictions are very close to ground truth in such cases. However, it is interesting to see that even with lower correlations ($0.2 \leq \rho <0.6$), the distribution of same percentile surgeon's is still on the higher side, especially for events based metrics. This means that a very high correlation between predicted and ground truth metrics may not be necessary when providing feedback in terms of performance quartiles.

Even with promising results, several limitations exist in this work. Our approach of primarily using visual information for model training performs well enough but has some short comings. Different tasks within a procedure can look very similar e.g. Anterior/Posterior bladder neck dissection, Posterior/Anterior anastomosis etc. The similarity in anatomy and the type of tools used can result in model getting confused between such tasks. In order to cater for this, methods to exploit the temporal nature of the overall procedure within the model can help reduce such mis-classifications. Moreover, while the LSTM part of the model does capture motion features from videos, it would be better to include other modalities like optical flow within the model for better motion feature learning. Furthermore, our approach does not take into account the problem of class imbalance. Since the number of images per task are not equal, methods to incorporate this information in the model training e.g. a weighted loss function, can further improve model performance.

Apart from model improvements, other parts of the pipeline can be improved. Firstly, our median filter approach is quite simple and likely contributed to inaccuracies in task boundary predictions. More advanced methods to post-process the output of the models should be explored, such as Hidden Markov Models and Conditional Random Fields.  Furthermore, these models should incorporate additional state information from the system in order to improve boundary estimation. One example could be which instruments were used for certain tasks \cite{Malpani2016}. Additional, multiple human annotators could provide better consensus of ground truth which is used to train different models.

\section{Conclusion}

In this paper, we showed that CNN-LSTM based models achieve the best performance for surgical activity recognition in robot-assisted prostatectomy and proposed a new, applied method to evaluate machine learning models based on their impact to efficiency metrics. Given the results presented in this paper, it appears feasible that post-operative efficiency reports can be automated, especially for a subset of critical tasks within an overall clinical procedure.

\noindent \textbf{Conflict of Interest:} 
Anthony Jarc, Linlin Zhou, and Liheng Guo are employees of Intuitive Surgical, Inc. Irfan Essa and Aneeq Zia declare that they have no conflict of interest.

\noindent \textbf{Ethical approval:} “All procedures performed in studies involving human participants were in accordance with the ethical standards of the institutional and/or national research committee and with the 1964 Helsinki declaration and its later amendments or comparable ethical standards.”

\noindent \textbf{Informed consent:} “Informed consent was obtained from all individual participants included in the study.

\bibliographystyle{splncs}
\bibliography{IJCARS}   
\end{document}